\documentclass[twocolumn]{article}

\everymath{\displaystyle}
\usepackage{amsmath,amsfonts,amssymb,amsthm}

\usepackage{url}

\usepackage{graphicx}
\usepackage[font=footnotesize,labelfont=bf]{caption, subcaption}
\usepackage{booktabs,tabularx}
\usepackage{float}
\usepackage{paralist}
\usepackage{xcolor}
\usepackage{algorithm,algorithmic}

\usepackage{gensymb}

\begin{document}

\title{Robust Assessment of Real-World Adversarial Examples}
\author{Brett Jefferson, Carlos Ortiz Marrero}

\maketitle
\begin{abstract}
    % We explore metrics to evaluate the robustness of a real-world adversarial attack, in particular adversarial patches, to changes in environmental conditions. We demonstrate how these metrics can be used to establish model baseline performance to then compare against real-world adversarial attacks. We establish a custom score for an adversarial condition that is adjusted for different environmental conditions and explore how the score changes with respect to specific environmental factors. Lastly, we propose two use cases for confidence distributions in each environmental condition. 
    We explore rigorous, systematic, and controlled experimental evaluation of adversarial examples in the real world and propose a testing regimen for evaluation of real world adversarial objects. We show that for small scene/ environmental perturbations, large adversarial performance differences exist. Current state of adversarial reporting exists largely as a frequency count over a dynamic collections of scenes. Our work underscores the need for either a more complete report or a score that incorporates scene changes and baseline performance for models and environments tested by adversarial developers. We put forth a score that attempts to address the above issues in a straight-forward exemplar application for multiple generated adversary examples. We contribute the following: \begin{inparaenum}\item a testbed for adversarial assessment, \item a score for adversarial examples, and \item a collection of additional evaluations on testbed data. \end{inparaenum}

\end{abstract}

%%%%%%%%%%%%%%%%%%%%%%%%%%%%%%%%%%%%%%%%%%%%%%%%%%%%%%%%%%%%%%%%%%%%%%%%%%%%%%%%%%%%%%%%%%%%%%%%%%%%
\section{Introduction}
Now that Deep Learning is an established success \cite{lecun2015}, there is a rapidly expanding body of work assessing its limitations \cite{szegedy2013, goodfellow2014, battaglia2018}. In particular, there has been a large number of papers published in recent years, interested in finding new ways to hack deep learning systems with a focus on manipulating convolutional neural networks into false and missed classifications \cite{papernot2016, moosavi2016, carliniweb2019}. Much of the early work with so-called adversarial attacks were only successful in virtual environments, {\it i.e.} the adversarial inputs were produced and evaluated digitally and without consideration of physical limitations. In the past year, researchers have expanded adversarial attacks to include physically created objects that can impact classifiers and detector models in real-world systems \cite{kurakin2016, athalye2017, eykholt2017, thys2019}. The range and ability of physical attacks are improving at an impressive rate, accounting for a variety of real-world considerations including static scenes, robust angle, and distance changes. %and video recorded from a moving vehicle. Reference
Although researchers have established some guidelines for evaluating the robustness of virtual adversarial attacks \cite{carlini2019}, the analysis recommendations do not map subjectively onto physical adversarial attacks where perturbations are difficult to measure and success varies on a frame by frame basis. The consistent computational measure of success for physical attacks is the percent of frames the attack accurately manipulated the classifier or detector \cite{eykholt2018, athalye2017, zhao2018}. This is separate from adversarial generation metrics that are often included in the optimization loss function to improve physical challenges like imperceptibility and printability. In our work we propose an evaluation experiment and post-generation, effectiveness score for testing the robustness of real-world adversarial examples in different environmental conditions. In particular, we tested our score on adversarial ``patches'', an idea clearly outlined in work done by Thys et al \cite{thys2019}, but incorporating ideas from Athalye et al. \cite{athalye2017}, Chen et al. \cite{chen2019}, and Eykholt et al. \cite{eykholt2018}. The key aspect we wish to address with our score is the importance of baseline performance across different environmental conditions when assessing the effectiveness of a given physical adverserial object.

% \begin{figure}[h!]
% 	\centering
% 	\includegraphics[width=\columnwidth]{Figures/frame-49.png}	
% 	\footnotesize
% 	\caption{YOLOv2: Real-Time Object Detector running in our lab.}
% 	\label{fig:YoloInLab}
% \end{figure}

%%%%%%%%%%%%%%%%%%%%%%%%%%%%%%%%%%%%%%%%%%%%%%%%%%%%%%%%%%%%%%%%%%%%%%%%%%%%%%%%%%%%%%%%%%%%%%%%%%%%
\section{Our Aim}
We recognize that it is not always possible to recreate environments to evaluate adversarial versus non-adversarial scenarios. However, many of the existing adversarial objects have the ability to be evaluated in well-controlled scenarios that can be reproduced.

We believe that in conjunction with well-controlled experiments, a proper adversarial score must be included when evaluating adversarial objects. Our proposed score does so in simple to interpret terms and is aimed at providing a foundation that can be updated, revised, and enhanced by others studying the problem of adversarial scoring. The score is designed to be a relative measure of performance that considers only those environments studied by the researcher while taking into account the same scenario in a non-adversarial condition. This is, in one sense, akin to something like a Bayesian information criterion score that only measures the model and parameters presented and cannot explicitly account for non-nested models. 

Experimentally, we studied a single target object with 3 differently-trained adversarial patches and an occlusion condition in several well-controlled scenarios. To the best of our knowledge, this is the most systematic assessment of adversarial attacks to date. For comparison, \cite{zhao2018} studied the effectiveness of adversarial attacks in an indoor environment and outdoor environment. While studying the attacks under similar scenes (across multiple days and weather conditions), by making changes in angle, distance, and day, the authors did not control for confounding factors such as other objects being present in the scene (scene complexity) nor did they report a baseline for the model. Our work aims at providing a coarse view of adversarial effectiveness using a fine-grained paradigm.

The paradigm in our work can be expanded to include more scenes/ environments, but we found it important to have an initial study that avoided confounding factors (such as patch performance being modulated by distance of the patch from the camera), while still providing plausible scene perturbations. For example, using our approach we can begin to quantify scene appropriateness or complexity, although this question deserves its own dedicated exploration.
%%%%%%%%%%%%%%%%%%%%%%%%%%%%%%%%%%%%%%%%%%%%%%%%%%%%%%%%%%%%%%%%%%%%%%%%%%%%%%%%%%%%%%%%%%%%%%%%%%%%
\section{Experiments}

Many researchers focus on the performance of an adversarial attack in native environmental conditions (e.g. a patch on the bumper of a vehicle in actual traffic, a patch attached to a stop sign, or a patch attached to a person's clothing in an office space). We assess performance in a well-controlled environment with little frame-to-frame variability due to moving objects, novel items entering the scene, changing lighting, etc. This tight control is necessary to accurately compare patch performance to a baseline condition where the model is allowed to detect objects without adversarial interference. In other words, we needed an environment that was reproducible and where happenstance occurrences were not a factor. For reproducibility, below we outline our experimental equipment and setup.

\subsection{Equipment}
Equipment for the experiment included scene setting items and camera hardware. Our scene setting items include a custom constructed mounting rail measuring roughly 7 feet long with attached platform (4 feet high) with plate for attaching camera devices and light source with one 40 watt, 390 lumens halogen bulb and one 40 watt, 450 lumens LED bulb. Camera and GPU devices include Jetson AGX Xavier, Jetson Xavier Developer Kit with an attached e-CAM130\_CUXVR camera. We tested 3 pre-printed adversarial patches created via different algorithm methods.

% \subsection{Additional Experimental Details}
% Adversary developers often do not report important object properties like dots-per-inch, printer color range, materials, glossiness or texture, how and where to affix adversarial objects with respect to target objects, effective sizes, or even ideal environmental conditions. 
% % This leaves much trial and error work up to researchers wishing to replicate attacks. 
% For our experiment, we conducted this trial and error work in our pilot where we tested various patch sizes, printing dpi settings, patch locations, lighting arrangements, and camera settings (focus, zoom, white balancing, exposure times, sharpness values, etc). The e-CAM130\_CUXVR camera we used allowed for manual setting of many of these aspects. Due to it's low focal length (2.8 mm), low f-number (2.8), and high field of view ($134\degree$), we were able to achieve relatively close distances for the patch and vase. Ultimately, our experimental setup was such that the patch was chosen to maximally hide (that is mis-classify or lead to a non-detection) the vase for most of the viewing conditions.

%%%%%%%%%%%%%%%%%%%%%%%%%%%%%%%%%%%%%%%%%%%%%%%%%%%%%%%%%%%%%%%%%%%%%%%%%%%%%%%%%%%%%%%%%%%%%%%%%%%%
\subsection{Patch Generation}\label{PatchGeneration}
We generate our patches using the training technique outlined by Thys et al. \cite{thys2019}. The broad idea can be summarized as follows:
\begin{enumerate}
    \item Curate a set of training images that your object detector can recognize.
    \item For each image we superimposed a patch ($300\times300$ pixels, then scaled accordingly to fit the size of the object bounding box) to the image. We use the Expectation over Transformation algorithm to produce our adversarial patch using the following transformations: change of location, rotation angle, scale, brightness, contrast, and noise level \cite{athalye2017}.
    \item We extract a classification score from these altered images, back-propagate the gradients to the input layer, and only update the pixels inside the region of the patch.
\end{enumerate}

We leverage and extend the code-base provided by Thys et al. \cite{thyscode} to generate adversarial patches for all classes contained in the output of our model. In our case, these are all classes in the COCO dataset \cite{lin2014}.

We trained three patches for the vase class (our target object) using images from ImageNet \cite{deng2009} and OpenImages \cite{kuznetsova2018}. Our patches were obtained by minimizing two different objective functions: objectness score (O) and the product of class probability score and objectness score (CxO). For our ImageNet patches we triage images from the WordnetID {\textit{n04522168}}, corresponding to the ImageNet ``vase'' class. For our Composite patch, we combined our extracted images from ImageNet with images extracted from OpenImages corresponding to the ``vase'' class name.

%%%%%%%%%%%%%%%%%%%%%%%%%%%%%%%%%%%%%%%%%%%%%%%%%%%%%%%%%%%%%%%%%%%%%%%%%%%%%%%%%%%%%%%%%%%%%%%%%%%%
\subsection{Procedure}

During piloting we found our patches should be $2 \times 2$ inches. We manipulated three environmental conditions/ dimensions: target item distance (1 inch, 5 inches, 10 inches, 15 inches and 20 inches from a Plexiglas surface), patch placement location (center and slight right of center), and lighting (2 bulb types). This brought the total number of environmental conditions to $5 \times 2 \times 2 = 20$. There were five patch conditions including no-patch (baseline), ImageNet (O), ImageNet (CxO), Composite (CxO), and a white patch, which was a simple white 2 inch square cut from 8.5 $\times$ 11 inches, 92 bright copy paper. All patches were printed on the aforementioned office paper at 1200 dpi. Section \ref{PatchGeneration} describes in detail how the patches where generated. Each patch type except for the no-patch condition was used in all environmental conditions. For the no-patch condition, the target item was always center-placed, but used with all light source and distance combinations. 

Our camera was manually set so that white balance, exposure time, and focus was fixed throughout the experiment (rather than automatically adjusted by the camera). The focus was set so that the sticker and vase were in focus from the camera (focal length: 2.8 mm; F-number: 2.8; Field of View: $134\degree$ (D), $73\degree$ (V)). We were able to position the camera at closer positions with such a wide field of view and shorter focal length. All other camera settings were constant throughout the data collection.

Our camera was placed 5 inches from a Plexiglas surface affixed to a table. The Plexiglas served as a mounting structure to ensure consistent placement for each patch. There was a single light source used at a time and the source was positioned behind and above the camera, pointed toward the target placement region. The camera was placed on the rail platform and fixed for the duration of the experiment. A large black board was used as background for the experiment. Testing proceeded as follows:

\begin{enumerate}	
    \item For a given distance, the target item (a green vase, see Figure \ref{fig: DoubleVase}) was placed in one of two positions depending on the patch placement condition. 
    \item With the target fixed, the no-patch condition was recorded first. Then each patch (white patch, ImageNet (O), ImageNet (CxO), and Composite (CxO)) was placed on the Plexiglas. Each time a patch was placed, it remained untouched through both lighting conditions. This was to minimize object and patch shifts across the conditions.
    \item Once a scene was established, we allowed 30 seconds for the bulb to warm-up.
    \item We ran a script that captured 500 frames and used each frame as an independent input to YOLOv2 \cite{redmon2016yolo9000}. We wanted enough frames for a single scene for a robust evaluation of that condition in lieu of natural image variation produced by the camera or by nature.
    \item We recorded bounding boxes, confidences, objectness scores for each frame.
\end{enumerate}

%%%%%%%%%%%%%%%%%%%%%%%%%%%%%%%%%%%%%%%%%%%%%%%%%%%%%%%%%%%%%%%%%%%%%%%%%%%%%%%%%%%%%%%%%%%%%%%%%%%%
\section{Results and Effect Score}

\begin{table*}[h!]
	\centering
	\footnotesize
	\caption{Number of Vase Detections (and Average Confidence) for Each Condition}
	\label{tab: N2}
	*When no patch was present, targets were center located. Thus, only one run of the model at different distances and bulb types was completed without a patch.
	\begin{tabular}{>{\bf }l >{\bf }l >{\bf }l|r|rrrr}
		\toprule
		Location & Bulb & Distance & None & Composite (CxO)   & ImageNet (CxO) & ImageNet (O) & White Patch \\
		\midrule
		center & Hlgn & 1 inch &	182 (.856) &	7 (.609) &	 0 &	 500 (.785) &	500 (.822) \\
		&	 & 5 inches &	404 (.905) &	 20 (.828) &	 0 &	500 (.869) &	500 (.887) \\
		&	 & 10 inches &	500 (.879) &	0 &	 0 &	 30 (.795) &	0 \\
		&	 & 15 inches &	298 (.846) &	0 &	 0 &	1 (.771) &	0 \\
		&	 & 20 inches &	499 (.823) &	0 &	 0 &	 32 (.670) &	0 \\
		
		& LED & 1 inch &	0 &	419 (.818) &	955 (.602) &	500 (.846) &	500 (.834) \\
		&	 & 5 inches &	500 (.928) &	500 (.892) &	391 (.691) &	500 (.938) &	500 (.915) \\
		&	 & 10 inches &	500 (.872) &	 27 (.870) &	 0 &	500 (.823) &	456 (.766) \\
		&	 & 15 inches &	500 (.872) &	0 &	 0 &	500 (.763) &	0 \\
		&	 & 20 inches &	500 (.880) &	0 &	 0 &	 28 (.694) &	0 \\
		
		right & Hlgn & 1 inch &	182* (.856) &	0 &	18 (.835) &	499 (.817) &	479 (.817) \\
		&	 & 5 inches &	404* (.905) &	0 &	206 (.914) &	500 (.904) &	313 (.899) \\
		&	 & 10 inches &	500* (.879) &	206 (.855) &	 0 &	302 (.819) &	1 (.715) \\
		&	 & 15 inches &	298* (.846) &	0 &	 0 &	 26 (.761) &	0 \\
		&	 & 20 inches &	499* (.823) &	0 &	 0 &	1 (.551) &	0 \\
		
		& LED & 1 inch &	0* &	2 (.820) &	457 (.851) &	500 (.867) &	500 (.866) \\
		&	 & 5 inches &	500* (.928) &	247 (.928) &	500 (.947) &	500 (.934) &	500 (.931) \\
		&	 & 10 inches &	500* (.872) &	500 (.893) &	500 (.859) &	500 (.840) &	500 (.807) \\
		&	 & 15 inches &	500* (.872) &	479 (.878) &	500 (.860) &	500 (.689) &	490 (.855) \\
		&	 & 20 inches &	500* (.880) &	0 &	 0 &	0 &	120 (.727) \\
		
		\bottomrule
	\end{tabular}
\end{table*}

\subsection{Classifications}

\begin{figure}[ht]
	\centering
	\footnotesize
	\caption{Patches Generated for Experiment}
		\begin{subfigure}{.25\columnwidth}
			\centering
			\includegraphics[width=\columnwidth]{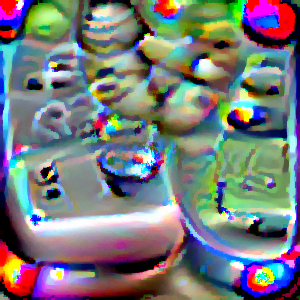}
		\end{subfigure}
		\begin{subfigure}{.25\columnwidth}
			\centering
			\includegraphics[width=\columnwidth]{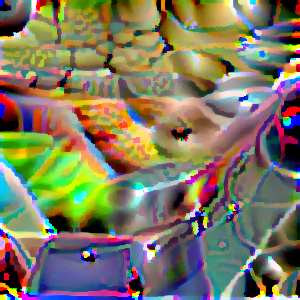}
		\end{subfigure}
		\begin{subfigure}{.25\columnwidth}
			\centering
			\includegraphics[width=\columnwidth]{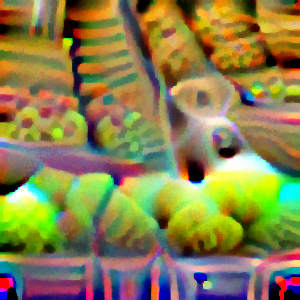}
		\end{subfigure}
\end{figure}

Each of the above patches was designed to hide a target item (i.e. a green vase) from detection for the YOLOv2 classifier. In addition to these patches a simple white square patch was also used to compare performance with an obstruction case. A first evaluation of each patch's ability to hide the target was to simply count the number of frames the classifier was able to detect the target class in each scene (See Table \ref{tab: N2}). This is a standard measure. Higher values in the table indicate the patch was not effective at disrupting detection of the target.

\begin{figure}[h!]
	\centering
	\includegraphics[width=.7\columnwidth]{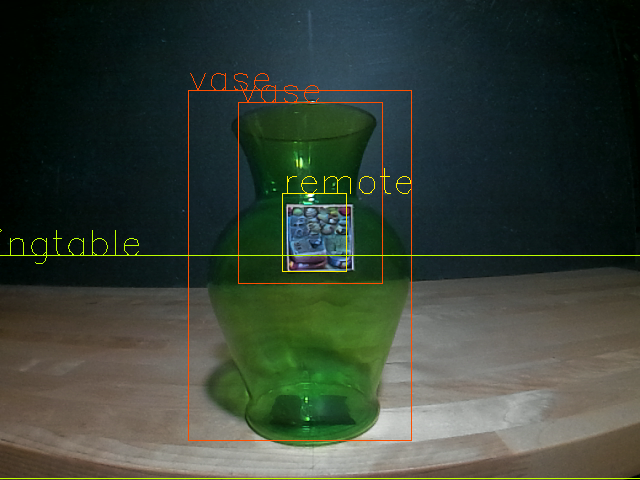}
	\caption{In this condition the vase was detected twice in nearly all frames.}
	\label{fig: DoubleVase}
\end{figure}

There are a few conditions that stand out when looking at only frequencies. When the target item was placed very close to the camera (1 inch condition) and \textbf{no patch was present}, the classifier had difficulty detecting it. In all frames with LED lighting, the target was missed, while in the halogen bulb lighting, the target was detected in less than 40\% of frames. Another stand-out is the LED, 1-inch, ImageNet (CxO) condition where the number of detections is higher than 500. In this case, the target is detected twice, once as a lower identification of the target and an upper identification of the target (see Figure \ref{fig: DoubleVase}). 

Judging from frequency alone, one might conclude that the Composite (CxO) patch and ImageNet (CxO) patch are \textit{better} than the other two patches. This conclusion would match intuition since one patch had a larger training set and both of these patches were trained using an objective function accounting for both class score and objectness. But there is more to be discovered. For instance, suppose one desired an all-around effective patch for a variety of physical environments. Is the Composite (CxO) patch better than the ImageNet (CxO) patch? A problem with error frequency is that is does not account for how well the model performs without adversarial interference. In the LED, 1-inch case, not having a patch at all is better than adding anything to the scene. We seek to develop a score that not only accounts for a variety of environmental changes, but also accounts for baseline performance in one summary.

\begin{figure}[h!]
	\centering
	\includegraphics[width=\columnwidth]{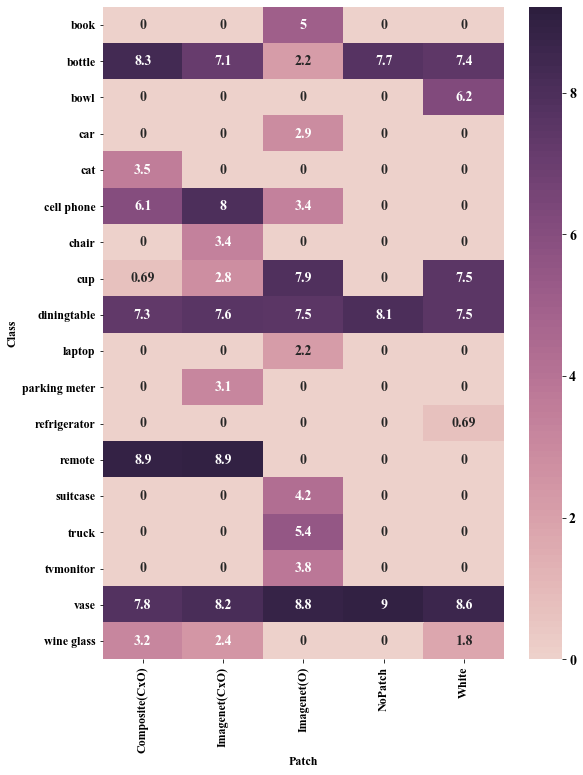}
	\caption{Depiction of all classifications per patch. These are summed over all environmental conditions and logged.}
	\label{fig: AllClass}
\end{figure}

Table \ref{fig: AllClass} shows the classifications per patch throughout the experiment. Because YOLOv2 is an object detector, multiple objects can be classified in a given frame. As a result, many classified objects are not misclassifications of the vase, but misclassifications of other scene objects. While, we attempted to minimize this effect, we often found that the table the vase was placed on was classified as a dining table and the background was classified as a refrigerator. More consistent misclassifications included classifying the vase as a bottle or a cup. `Bottle' labels occurred frequently, even in the absence of an adversarial patch. The patches themselves were classified in some many instances. The patches trained to decrease Class Score and Objectness were classified as a cell phone or remote, while the third patch (optimized for Objectness only) was classified as a wider range of objects. Here we reemphasize that no patch hid the vase 100\% of the time, but that there are some scenes where patches performed well and others that it simply did not work. We did a parameter sweep to pick penalties for the non-printability and total variation term in the loss function and trained all patches until we saw no improvement in the loss function. The question of producing the ``optimal'' patch was outside the scope of our work, given that our main focus was assessing adversarial patches.

%%%%%%%%%%%%%%%%%%%%%%%%%%%%%%%%%%%%%%%%%%%%%%%%%%%%%%%%%%%%%%%%%%%%%%%%%%%%%%%%%%%%%%%%%%%%%%%%%%%%%%%%%
%%%%%%%%%%%%%%%%%%%%%%%%%%%%%%%%%%%%%%%%%%%%%%%%%%%%%%%%%%%%%%%%%%%%%%%%%%%%%%%%%%%%%%%%%%%%%%%%%%%%%%%%%
\subsection{Effectiveness Score}

To gain a better understanding of patch performance, we make the straightforward adjustment of comparing patch performance for a given scene with model performance in the absence of an adversary. As noted above, there were several misclassifications of the vase in the no patch condition, leading to potential misunderstanding of the effectiveness of a given adversary. Our proposed score is derived for each patch in a given scene/ condition (or conditions). For a given patch $P$ and a single environmental condition $e$ (e.g. lighting, distance, patch location), we compute the frequency of  target detection in both an adversary condition (patch present) and baseline (adversary not present). Let the $f_{P,e}$ be the frequency for a given patch and environmental condition (out of the set of tested environmental conditions) $e \in E$. We let $f_{\emptyset, e}$ denote the frequency of  target detection of baseline. Let $n$ denote the total number of frames captured. We define the score for a patch conditionally over the set of environments to be
\begin{align*}
	S(P, E) & = \dfrac{1}{|E|}\sum_{e\in E} \frac{\left(f_{\emptyset, e} - f_{P,e}\right)}{n} \\
\end{align*}
 This is the simply the average difference of detection probabilities between the no-patch and patch conditions. For this score function, when the target is detected in all 500 frames of our experiment and the patch successfully hides the target in all 500 frames, the score will be 1. When the baseline model is unsuccessful at detecting the target, the score is lower. Negative scores can be interpreted as the adversary having the highly-undesired effect of helping the model to detect a target more often, instead of less. Lastly, the score is averaged across all conditions run to provide a simple summary of performance. 
 
The scores for our patches are given in Table \ref{tab: PatchScores}. A current limitation is that the score cannot distinguish between poor baseline versus poor adversary.

 %It is interesting to note that by score, the white patch outscores the patch trained using only the ImageNet corpus with objective function only utilizing the \textit{objectness} score from YOLOv2.

\begin{table}[h!]
	\caption{Patch Scores}
	\label{tab: PatchScores}
	\centering
	\footnotesize
	\begin{tabular}{lr}
		\toprule
		Patch Condition & Score\\
		\hline
		No-Patch	& 0\\
		\textbf{Composite (CxO)}	& \textbf{0.536}\\
		ImageNet (CxO)	&0.424\\
		ImageNet (O)	&0.135\\
		White Patch	&0.241\\
		\bottomrule
	\end{tabular}
\end{table}
We also computed scores for each condition independent of the other conditions (each $E$ is a singleton). 
% We ran an ANOVA on this data with factors \textit{Patch Location}, \textit{Bulb Type}, \textit{Target Distance}, and \textit{Patch}. Bulb type, target distance, and patch were all significant factors.
The analysis in the next section dives into the pattern of results on these dimensions. 
% \begin{table}
% 	\centering
% 	\caption{ANOVA on Scores}
% 	\label{tab: ANOVA}
% 	\footnotesize
% 	\begin{tabular}{lrrrrr}
% 		\toprule
% 		&    df &     sum.sq &   mean.sq &          F &        PR($>$F)\\
% 		\hline
% 		Lctn.   &   1.0 &   0.16 &  0.16 &   1.04 &  3.11e-01\\
% 		Bulb    &   1.0 &   2.25 &  2.25 &  14.89 &  2.16e-04\\
% 		Dist.   &   4.0 &  16.43 &  4.11 &  27.21 &  9.31e-15\\
% 		Patch   &   4.0 &   3.73 &  0.93 &   6.18 &  1.97e-04\\
% 		Resid.  &  89.0 &  13.43 &  0.15 &        NaN &           NaN\\
% 		\bottomrule
% 	\end{tabular}
% \end{table}

%%%%%%%%%%%%%%%%%%%%%%%%%%%%%%%%%%%%%%%%%%%%%%%%%%%%%%%%%%%%%%%%%%%%%%%%%%%%%%%%%%%%%%%%%%%%%%%%%%%%%%%%%
%%%%%%%%%%%%%%%%%%%%%%%%%%%%%%%%%%%%%%%%%%%%%%%%%%%%%%%%%%%%%%%%%%%%%%%%%%%%%%%%%%%%%%%%%%%%%%%%%%%%%%%%%
\subsection{Dimension Impacts}

Recall that for a 1 inch distance, baseline model performance was particularly poor. However in the ImageNet and white patch conditions model performance \textbf{increased} significantly, regardless of lighting or center/ right patch placement. This may indicate that pre-trained YOLOv2 is not robust to large-scaled objects. Somehow, more \textit{generic} occlusions provides YOLOv2 with enough context to make an accurate identification.

\begin{figure}[h!]
	\centering
	\footnotesize
	\caption{Image capture of YOLOv2 detection at 5 distances with white patch under LED lighting.}
	\label{fig: DistImages}
	\begin{subfigure}{.30\columnwidth}
		\includegraphics[width = \columnwidth]{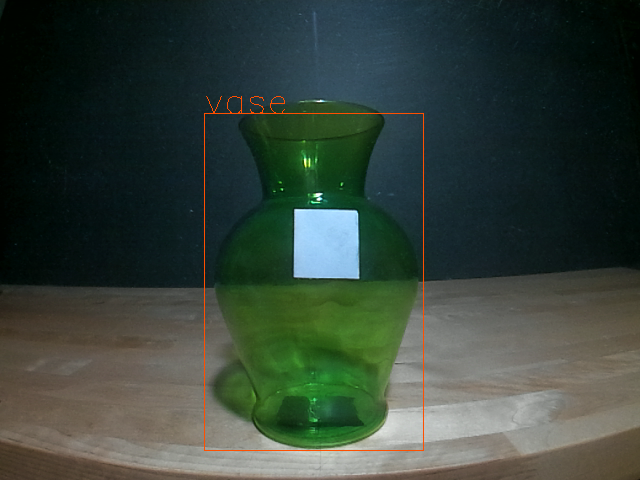}
	\end{subfigure}
	\begin{subfigure}{.30\columnwidth}
		\includegraphics[width = \columnwidth]{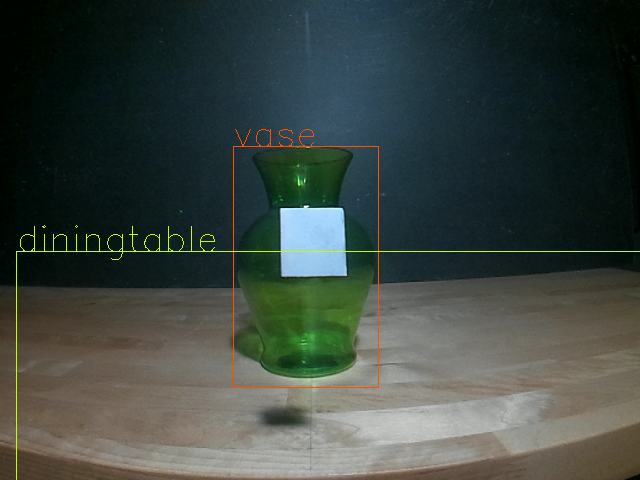}
	\end{subfigure}
	\begin{subfigure}{.30\columnwidth}
		\includegraphics[width = \columnwidth]{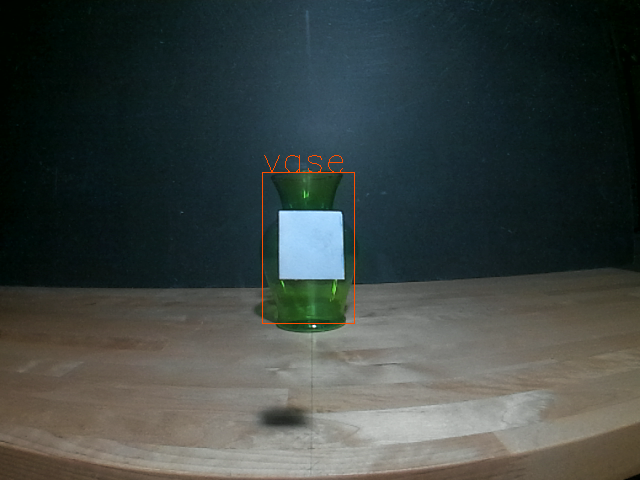}
	\end{subfigure}
	\begin{subfigure}{.30\columnwidth}
		\includegraphics[width = \columnwidth]{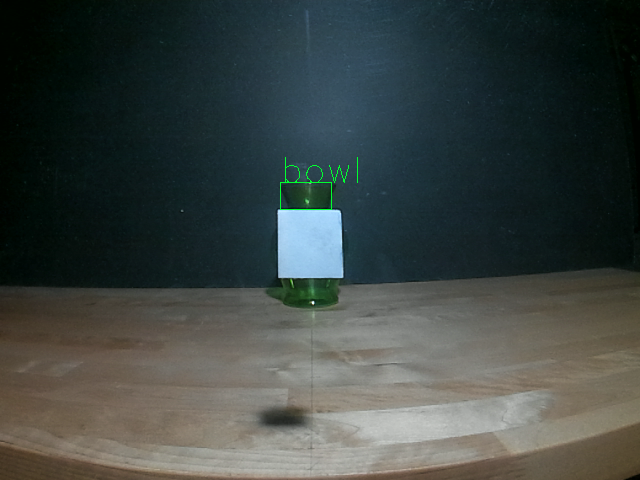}
	\end{subfigure}
	\begin{subfigure}{.30\columnwidth}
		\includegraphics[width = \columnwidth]{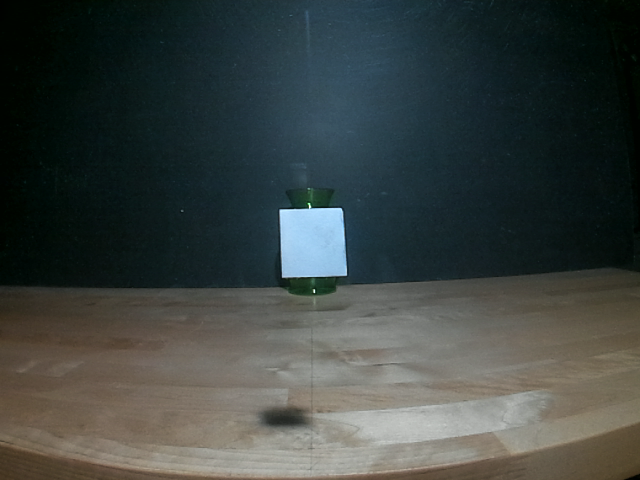}
	\end{subfigure}
\end{figure}

In addition to poor model/ patch performance for close distances, the scores dip when the target item is 15 inches from the Plexiglas. The dip could be driven by either low baseline model performance at 15 inches, or by poor patch performance. A quick look at Table \ref{tab: N2} reveals that baseline model performance also decreases at 15 inches (number of correct detections without the patch is 298 out of the possible 500 for halogen bulbs). When this occurs, the model score decreases since the score is relative to baseline. 

\begin{figure}[h!]
    \centering
	\caption{Mean Confidence per Distance}
	\label{fig: MeanbyDist}
	\includegraphics[trim=0 30 0 80, clip, width = .75\columnwidth]{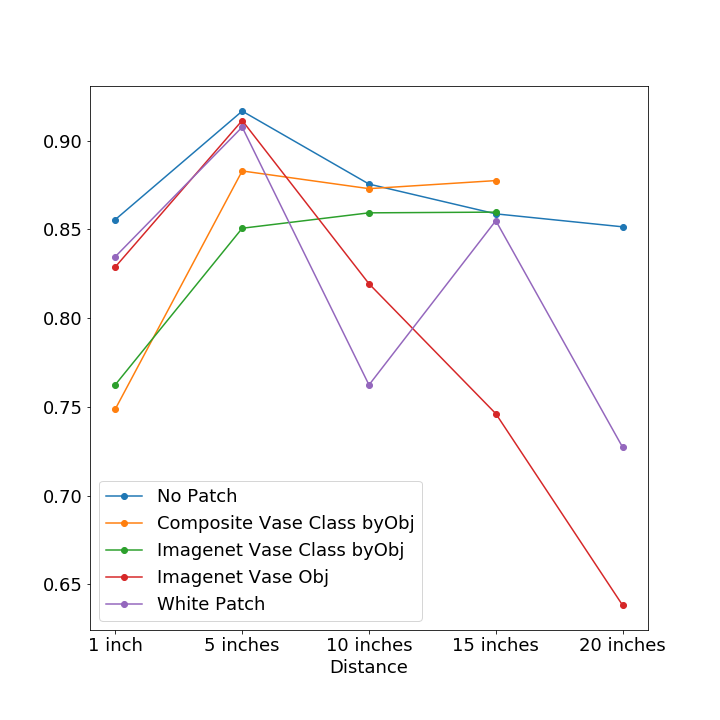}
\end{figure}

We also recorded confidence scores for each condition to study if there was a relationship with patch performance. Confidence scores did not influence patch performance across the distances. Figure \ref{fig: MeanbyDist} displays the average confidence for each distance and each patch condition. Baseline and ImageNet are the only two patch conditions that decrease from 5 inches to 20 inches. The two other simulated patches are consistently high for 5 to 15 inches and then effectively hide the target at 20 inches, while the white patch telescopes in performance. One might expect lower confidence scores leading to fewer detection. However, confidence scores are constructed independently of probability of detection within YOLOv2. The model detects a target when the objectness score is above 0.5 and to prevent multiple detections of the same object, the NMS threshold is set to 0.4. The two patch conditions with consistently high confidences when the target was detected are also the two models that are unable to detect the target the most. At 15 inches (where all simulated patches have a sudden decrease in score), the highest scoring patch, Composite (CxO), has a higher confidence when the target is detected than the baseline model. This provides some experimental evidence that confidence score alone and without context to other class scores are not a clear predictor of success for patches designed to hide a target.

\begin{figure}[h!]
	\centering
	\footnotesize
	\caption{\textbf{[Score by Distance for Each Patch]} Points are averaged values for each distance.}
	\label{fig: ScorebyDist}
	\includegraphics[trim=0 15 0 80, clip, width = .75\columnwidth]{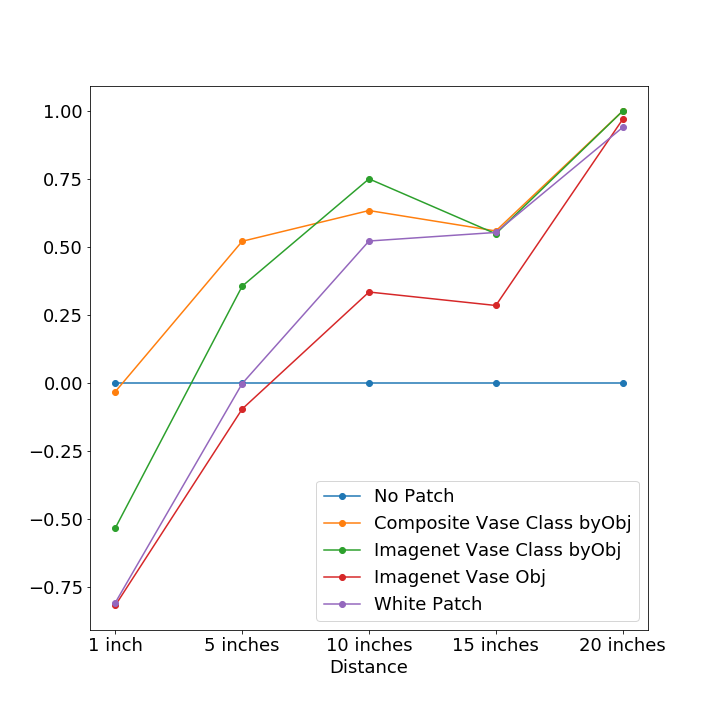}
\end{figure}

An alternative explanation for model performance beyond 10 inches is that the target itself is mostly occluded from camera view at further distances. More than 97\% of all frames at this distance were successfully either misclassified or not identified considering all patches. The sub-condition with the highest detection frequency occurred with the white patch placed slightly right of the target under LED lighting. Further testing is required to confirm occlusion is the main reason for poor performance. A counter condition is the ImageNet patch. When that patch was used there were 60 correct detections across lighting at center location, but only a single correct detection when the target was slightly right of the patch. The white patch, LED, right-position condition led to 120 of the 500 frames having correct detections at 20 inches.

% \begin{figure}[h!]
% 	\centering
% 	\includegraphics[width=.7\columnwidth]{Figures/CPCO20iHalCenter.png}
% 	\caption{In this condition the vase was misclassified in all frames.}
% 	\label{fig: 20inchHidden}
% \end{figure}

%%%%%%%%%%%%%%%%%%%%%%%%%%%%%%%%%%%%%%%%%%%%%%%%%%%%%%%%%%%%%%%%%%%%%%%%%%%%%%%%%%%%%%%%%%%%%%%%%%%%%%%%%
%%%%%%%%%%%%%%%%%%%%%%%%%%%%%%%%%%%%%%%%%%%%%%%%%%%%%%%%%%%%%%%%%%%%%%%%%%%%%%%%%%%%%%%%%%%%%%%%%%%%%%%%%
% Lighting

\begin{figure}[h!]
	\centering
	\footnotesize
	\caption{Image capture of YOLOv2 detection at 2 light configurations with white patch at 10 inches.}
	\label{fig: LightImages}
	\begin{subfigure}{.40\columnwidth}
		\includegraphics[width = \columnwidth]{Figures/10inch.png}
		\caption{LED Lighting}
	\end{subfigure}
	\begin{subfigure}{.40\columnwidth}
		\includegraphics[width = \columnwidth]{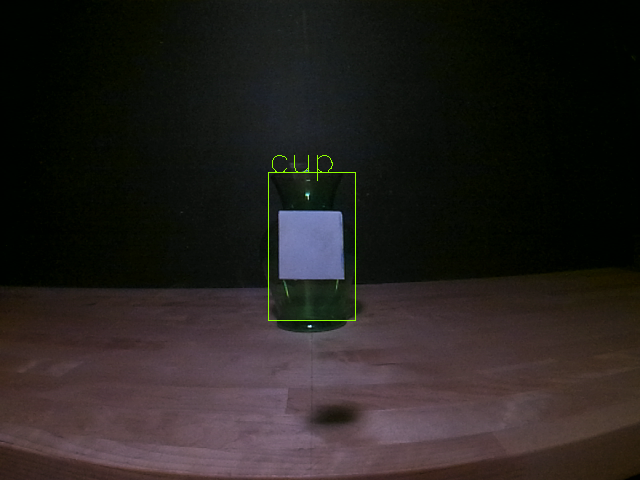}
		\caption{Halogen Lighting}
	\end{subfigure}
\end{figure}

Lighting was also a significant factor for patch scores. Figure \ref{fig: LightImages} displays the difference in lighting used for this experiment. We predicted that for the LED condition, the patches would be more effective at hiding the target. However, the LED condition resulted in fewer disappearances (more correct detections) than the halogen condition. This is surprising given the LED is more luminescent.

Considering only changes in this factor, we find that across the two lighting conditions, the Composite (CxO) and ImageNet (CxO) patches outperform the other two patches. In addition, the white patch has higher scores than the ImageNet (O) patch in both lighting conditions. Performance in the halogen bulb condition is limited to 0.7 by the fact that we are averaging scores across the other dimensions (location and distance) and there are cases in which the baseline model had missed detection occurrences in these conditions.

\begin{figure}[h!]
	\centering
	\footnotesize
	\caption{\textbf{[Score by Light for Each Patch]} Points are averaged values for each lighting source.}
	\label{fig: ScorebyLight}
	\includegraphics[trim=0 30 0 80, clip, width = .75\columnwidth]{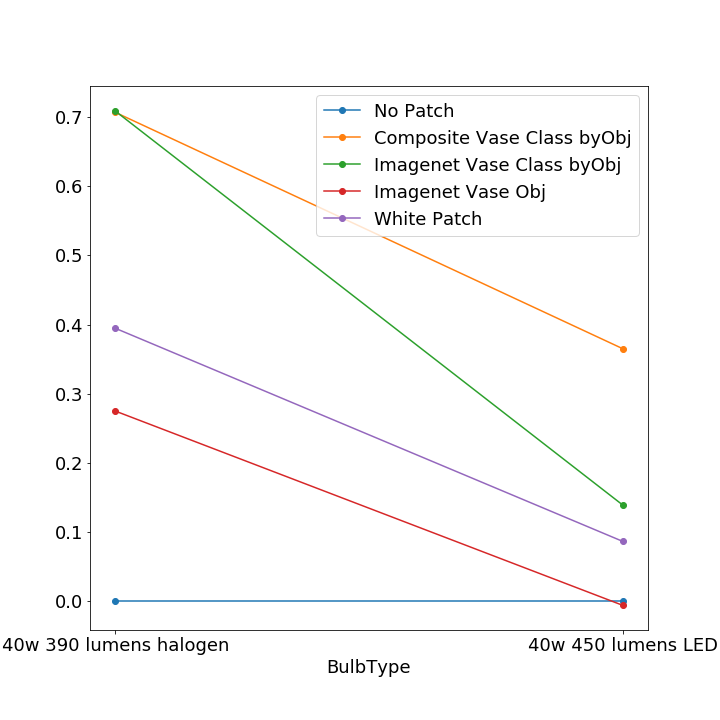}
\end{figure}

%%%%%%%%%%%%%%%%%%%%%%%%%%%%%%%%%%%%%%%%%%%%%%%%%%%%%%%%%%%%%%%%%%%%%%%%%%%%%%%%%%%%%%%%%%%%%%%%%%%%%%%%%
%%%%%%%%%%%%%%%%%%%%%%%%%%%%%%%%%%%%%%%%%%%%%%%%%%%%%%%%%%%%%%%%%%%%%%%%%%%%%%%%%%%%%%%%%%%%%%%%%%%%%%%%%
% Location

The same trend, although on a smaller scale, occurs when marginalizing over location values. 
% Figure \ref{fig: ScorebyLocation} depicts mean values at each location. 
We computed 68\% (roughly two standard deviations) confidence intervals by a bootstrapping procedure. Across the two target locations, the confidence intervals have high overlap which is an indicator that performance is likely equal regardless of target location. The Composite (CxO) and ImageNet (CxO) patches are the only two that had confidence intervals that did not overlap with a score of 0, indicating that for both location conditions, these patches had some affect on YOLOv2. However, after running one-way ANOVA's, even these patch scores were not significantly different from 0 (p =[0.323, 0.219, 0.969, 0.595] for Composite (CxO), ImageNet (CxO), ImageNet, and white patch respectively).

\begin{figure}[h!]
	\centering
	\footnotesize
	\caption{Image capture of YOLOv2 detection at 2 location configurations (center and right) with white patch at 10 inches under LED lighting.}
	\label{fig: LocationImages}
	\begin{subfigure}{.40\columnwidth}
		\includegraphics[width = \columnwidth]{Figures/10inch.png}
	\end{subfigure}
	\begin{subfigure}{.40\columnwidth}
		\includegraphics[width = \columnwidth]{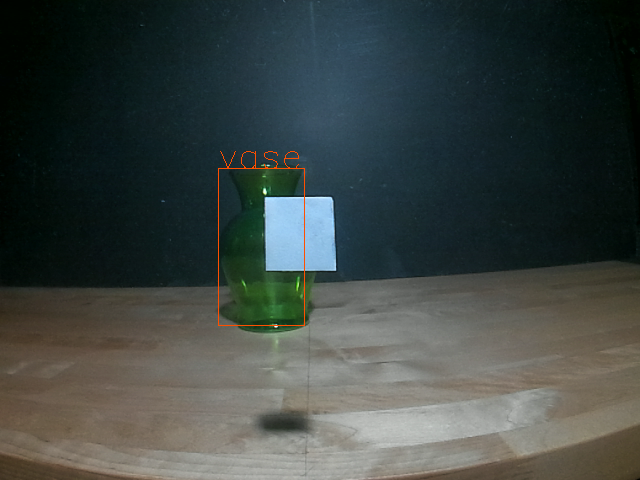}
	\end{subfigure}
\end{figure}

\section{Conclusion and Future Work}
This paper makes two contributions. First, we propose a score for adversarial attacks in the physical world. The score compares attack performance to a baseline. We compute the score in a controlled environment with reproducible environmental conditions. We include two potential use cases for the proposed score. The second contribution is that, to the best of our knowledge,  this is the most systematic assessment of adversarial attacks to date. Chen et al. \cite{chen2019} had an indoor systematic assessment of sticker attacks but did not investigate varying controlled lighting and placement, rather distance and angle. While many of the current papers highlight that their methods and patch generation methods work well in the real world, it is of importance to account for the weaknesses in any method to not only prevent other researchers from making the same mistakes, but to advance scientific understanding of deep learning models in general. Moreover, we can empirically conclude that camera aspects and model training are interacting with environmental conditions to produce odd model results (such as the baseline model not detecting a vase five and fifteen inches in front of the camera). Our aim with the approach taken was to strive towards a full-report of both the model and the adversarial object in the real-world and to highlight in detail the challenges researchers face when evaluating adversarial objects.

%\section*{Acknowledgement}
%The authors would like to thank Andrew August for generating the teapots and initial parsing of data.

% We should not give away next steps
% In future work, we wish to extend the score derived here to study whether there are any systematic patterns to model.
%and adversarial attack weaknesses. CANT SAY THIS
%One may notice that our investigation, 
%while exposing weaknesses in the YOLOv2 model with respect to training adversarial attacks, 
%does not provide an answer as to why the models are vulnerable to certain real-world factors. This is a difficult problem that many recent advances have attempted to solve. 
% WE REALLY DONT WANNA SAY THIS
%A good summary on this work is \cite{chak2017}. 

%%%%%%%%%%%%%%%%%%%%%%%%%%%%%%%%%%%%%%%%%%%%%%%%%%%%%%%%%%%%%%%%%%%%%%%%%%%%%%%%%%%%%%%%%%%%%%%%%%%%
\bibliographystyle{ieeetr} 
\bibliography{references}

\end{document}